\documentclass[11pt]{article}

\usepackage[final]{acl}
\usepackage{times}
\usepackage{latexsym}
\usepackage[T1]{fontenc}
\usepackage[utf8]{inputenc}
\usepackage{microtype}
\usepackage{inconsolata}
\usepackage{amsmath}
\usepackage{booktabs}
\usepackage{array}
\usepackage{graphicx}
\usepackage{multirow}
\usepackage{url}
\usepackage{tabularx}
\usepackage{adjustbox}

\newcommand{\benchmark}{\textsc{DelistBench}}
\newcommand{\framework}{\textsc{Search-to-Record}}
\newcommand{\yes}{\texttt{yes}}
\newcommand{\no}{\texttt{no}}
\newcommand{\unknown}{\texttt{unknown}}
\newcommand{\pp}{\,pp}

\title{\benchmark: Evaluating Search-Enabled LLMs for\\
Auditable Corporate-Event Database Completion}

\author{
  Xuan Yao \quad Shuping Li \quad Yang Dai \quad Yi Zhou \quad Ke-Wei Huang \\
  Asian Institute of Digital Finance, National University of Singapore, Singapore \\
  \texttt{\{yaoxuan, aidf.li, koreydai, zhouyi02, dishkw\}@nus.edu.sg}
}

\begin{document}
\maketitle

\begin{abstract}
Financial institutions need an independent way to detect missing, stale, and misclassified corporate-event records in vendor databases. We introduce \framework, a database-assurance task in which search-enabled large language models reconstruct institution-defined event records from public sources for a known security universe and historical cutoff, and \benchmark, a 1,200-record benchmark for security-level delisting announcements. We evaluate five models in paired closed-book and web-enabled conditions. Web access raises announcement-date accuracy within seven days by 34.0--48.0 percentage points and event-status accuracy by approximately 2.8--21.7 points; the best system achieves 81.5\% overall joint accuracy within seven days. Economy web systems achieve 75.9--78.3\% overall joint accuracy within seven days at 4.5--6.6\% of the API cost of the most expensive web system. Risk-based triage identifies low-error subsets, although the highest-coverage operating point still sends 27.3\% of the balanced test set to review. The evaluation identifies web retrieval as the main source of timing gains and shows that low-cost systems can approach the best system's accuracy. Together, \framework, \benchmark, and the evaluation provide concrete deployment guidance: calibrate triage to local event prevalence and market mix, preserve positive-event recall, and route positive and ambiguous cases to targeted review. \benchmark{} is publicly available at \url{https://github.com/tj-zyyyyyyy/DelistBench}.
\end{abstract}

\section{Introduction}

Financial institutions maintain a security master that identifies the issuers, securities, listings, exchanges, and identifiers in scope. Corporate-event histories attached to that universe are often purchased from global data vendors. The central operational uncertainty is therefore not usually \emph{which companies exist}; it is whether the event database is complete, accurate, timely, and consistent with the institution's definition. Coverage gaps may be concentrated in small markets, local-language disclosures, cross-listed securities, or event types whose procedural boundaries differ across exchanges.

An independent public-source audit is expensive. For every target listing, an analyst may need to locate exchange and issuer notices, distinguish a formal action from a warning, identify the announcement rather than effective date, map the stated reason, and retain an evidence trail. Search-enabled large language models (LLMs) can perform these steps at low API cost, but fluent output is not necessarily database-ready. A model can retrieve the correct company but wrong listing, project a later event backward across a historical cutoff, or return a plausible date unsupported by its citations.

We study this setting as \emph{closed-universe, open-event-status} completion. The target identity and cutoff are known; the existence and contents of a qualifying event record are not. We instantiate the task with delisting announcements, whose security-level scope, heterogeneous reasons, and easily confused procedural dates create a demanding test of retrieval and temporal adjudication.

Our contributions are:
\begin{itemize}
    \item We formulate \framework{} (S2R), an auditable database-assurance task that reconstructs institution-defined event records independently of a vendor feed.
    \item We construct and will publicly release \benchmark, a 1,200-record benchmark with balanced positive, later-event, and still-listed strata, public evidence, and issuer-family-disjoint splits.
    \item We compare five model families in paired closed-book and web-enabled conditions across accuracy, evidence, cost, and latency, and evaluate risk-based acceptance under both the balanced benchmark and a low-prevalence deployment framing.
\end{itemize}

\section{Related Work}

\paragraph{Financial event extraction.}
Financial event extraction typically assumes a supplied document. DCFEE extracts document-level events \citep{yang2018dcfee}; fine-grained resources annotate business-news triggers and arguments \citep{jacobs2020economic}; and FinReason targets reasons in company announcements \citep{chen2021finreason}. S2R instead includes retrieval and listing resolution and returns a database record rather than an event mention.

\paragraph{Search-grounded language models.}
WebGPT and ReAct established browser-assisted answering and interleaved tool use \citep{nakano2022webgpt,yao2023react}. FreshLLMs and RealTime QA study retrieval for changing facts \citep{vu2024freshllms,kasai2023realtimeqa}; financial RAG benchmarks instead provide a bounded document collection \citep{lithgow2025findoc}. Recent financial deep-research benchmarks study corporate analysis and forecasting \citep{zhu2025findeepresearch,li2026findeepforecast}. We use the open web but fix the target identity, event definition, and historical cutoff.

\paragraph{Evidence and selective prediction.}
ALCE evaluates citation quality \citep{gao2023alce}, while LongFact decomposes claims for search-based verification \citep{wei2024longfact}. Selective classification trades coverage for lower error \citep{geifman2017selective}; risk-control procedures can provide finite-sample guarantees under explicit assumptions \citep{bates2021risk}. Our calibrated-acceptance threshold enforces an empirical calibration-set criterion, not a formal test-set guarantee.

\section{Methodology}

\subsection{Search-to-Record Task}

\begin{figure*}[t]
    \centering
    \includegraphics[
        width=1\textwidth
    ]{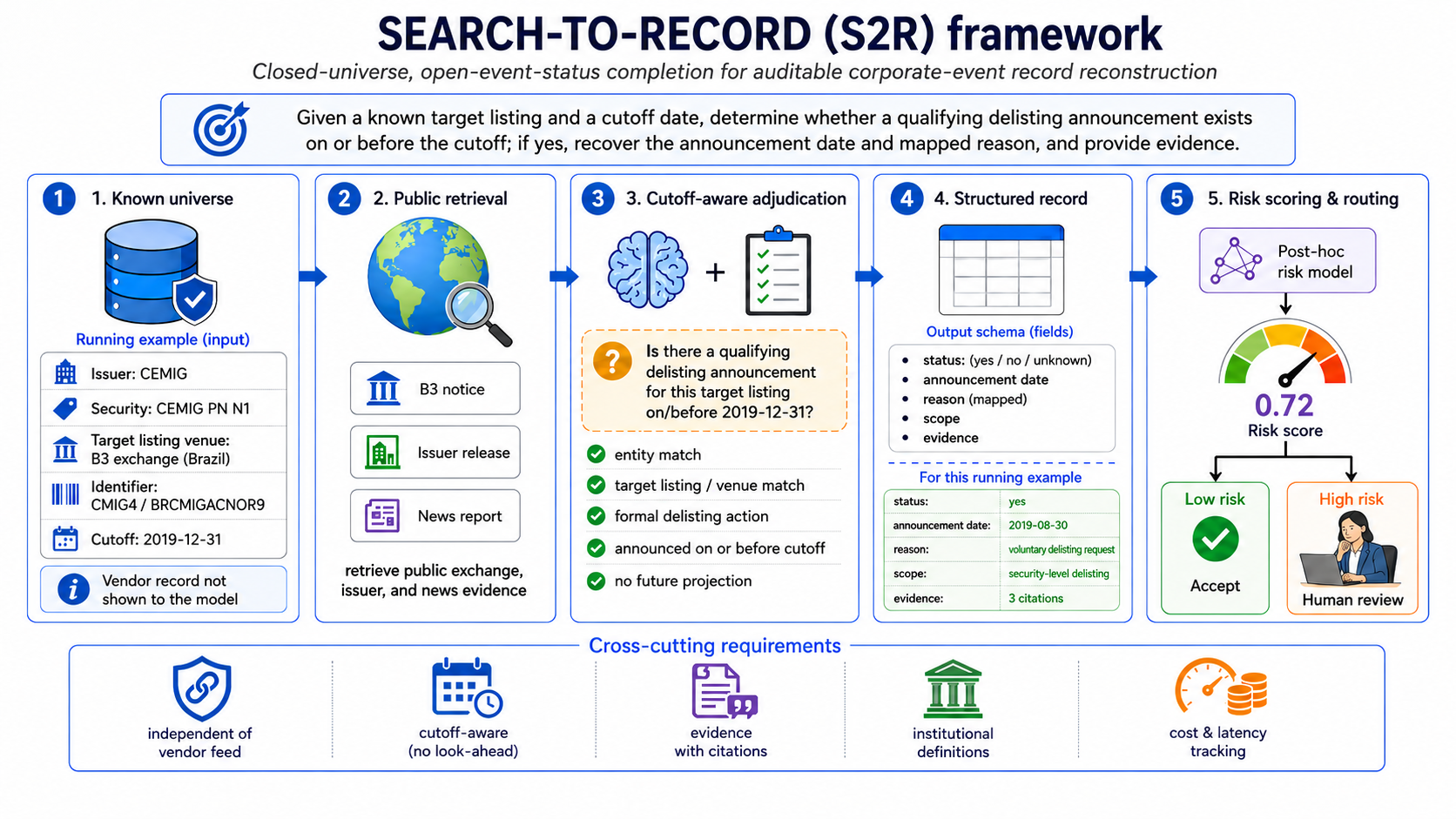}

    \vspace{-1mm}
    \setlength{\tabcolsep}{2pt}

    \caption{The five-stage \framework{} workflow: retrieve public evidence for a known target, apply cutoff-aware adjudication, construct a record, and route it to acceptance or review.}
    \label{fig:s2r}
\end{figure*}

An S2R input is
\begin{equation}
 x_i=(e_i,s_i,v_i,c_i,z_i,T_i),
\end{equation}
where $e_i$ is the issuer, $s_i$ the security, $v_i$ the target venue, $c_i$ the country, $z_i$ an optional stable identifier, and $T_i$ the cutoff. The system returns
\begin{equation}
 \hat y_i=(q_i,d_i,r_i,E_i,u_i),
\end{equation}
where $q_i\in\{\yes,\no,\unknown\}$ is status at the cutoff, $d_i$ the formal announcement date, $r_i$ a normalized reason, $E_i$ claim-level evidence, and $u_i$ uncertainty. Date and reason are null unless $q_i=\yes$.

The same output supports several database-assurance decisions: a missing vendor event, a stale update, an incorrect date or reason, an event attached to the wrong listing, or confirmation that no qualifying event existed by the cutoff. The vendor record is not shown to the evaluated model; public reconstruction therefore provides an independent comparison rather than a paraphrase of the incumbent feed.

\paragraph{Delisting event definition.}

A positive requires a publicly disclosed formal decision, approved application, exchange determination, filed withdrawal, or completed delisting action directed at the specified security and venue. A deficiency warning, cure period, possible delisting, suspension, last-trading date, effective date without a qualifying disclosure, or action concerning another listing is insufficient. Later reversal or continued trading does not invalidate a qualifying historical action. Current pages may reconstruct history, but the canonical date is the first qualifying public disclosure identified during benchmark construction, not the page-publication or effective-removal date. This makes later-event projection a first-class error rather than treating all retrieved delisting content as relevant.

\subsection{DelistBench}

\begin{figure*}[t]
    \centering
    \includegraphics[
        width=\textwidth
    ]{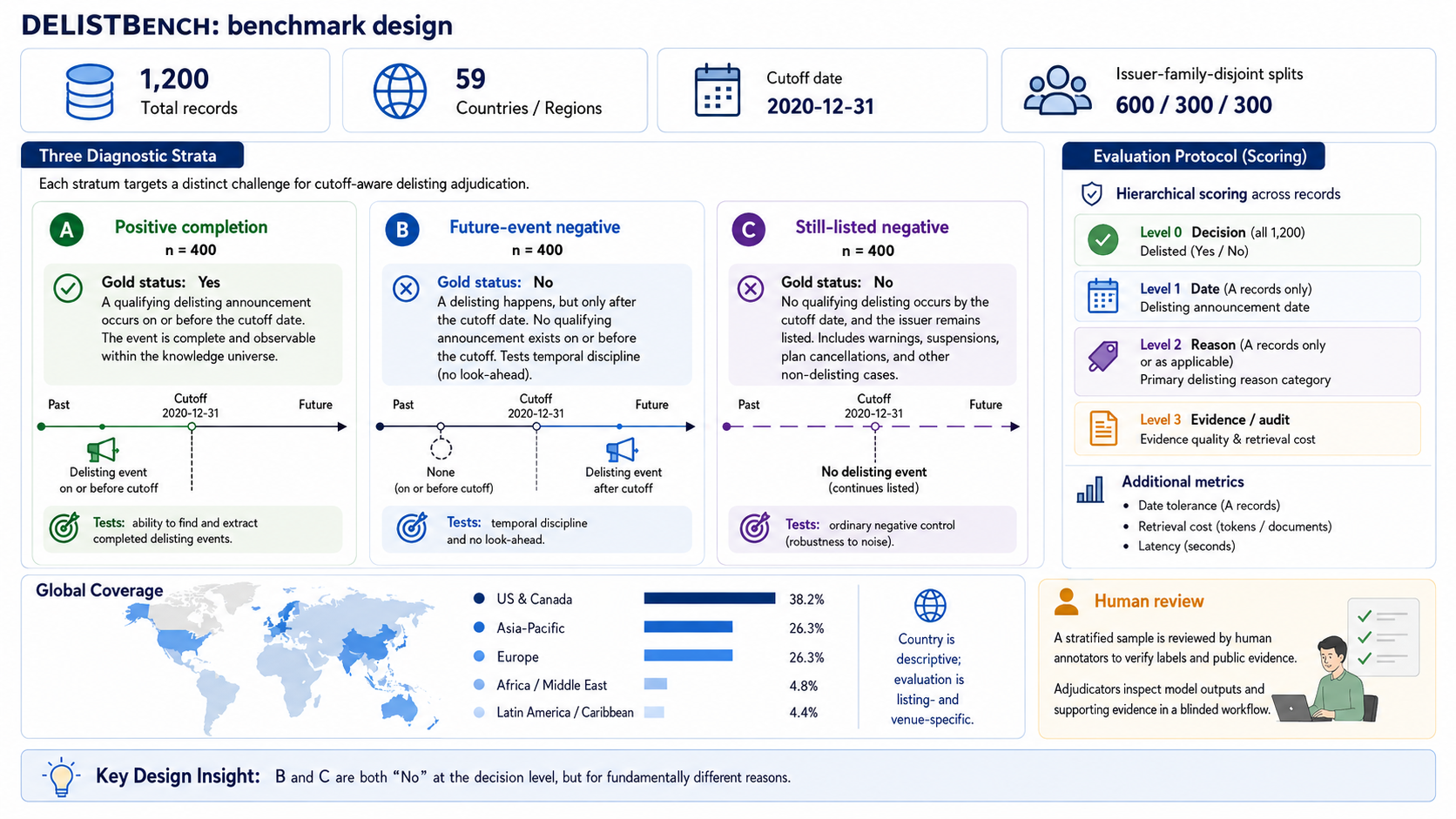}
    \caption{Overview of the \benchmark{} design and evaluation protocol.
    The three balanced strata distinguish positive completion (A) from
    future-event negatives (B) and still-listed negatives (C).
    Levels 0--2 define hierarchical record correctness, while evidence
    quality, retrieval cost, latency, and review requirements are evaluated
    separately. The 59-country count describes the announcement-candidate
    pool, and the human-review panel includes an independent
    reliability check. Evaluation remains listing- and venue-specific.}
    \label{fig:delistbench-design}
\end{figure*}

We instantiate \framework{} as cutoff-aware reconstruction of security-level
delisting announcements. Figure~\ref{fig:delistbench-design} summarizes the
resulting \benchmark{} benchmark and evaluation protocol.

\paragraph{Benchmark construction and diagnostic strata.}

\benchmark{} contains 1,200 records divided evenly among three diagnostic
strata, as illustrated in Figure~\ref{fig:delistbench-design}. Stratum A
contains 400 positive-completion records for which a qualifying delisting
announcement occurred on or before the cutoff. Stratum B contains 400 future-announcement negatives: no qualifying formal delisting action had been disclosed by the cutoff, but a qualifying announcement was issued only after the cutoff. Stratum C contains 400 active controls for which no qualifying formal delisting action was identified for the exact listing at any time through the observation date of 6 August 2026. Although B and C both have
gold status \no{}, they represent fundamentally different failure modes.

The announcement candidates span 59 issuer countries. Of the full benchmark, 38.2\% are from
the United States and Canada, 26.3\% from Asia--Pacific, 26.3\% from Europe,
4.8\% from Africa and the Middle East, and 4.4\% from Latin America and the
Caribbean. Country is descriptive; evaluation remains listing- and
venue-specific. Candidate A/B events originate from a licensed
corporate-event source used only for discovery and internal audit. Release
labels are reconstructed from permissible public evidence, and licensed text
is not sent to evaluated APIs. Category C records are matched on geography, industry, name shape, and approximate difficulty. No qualifying formal delisting action was identified for the exact listing through 6 August 2026, and the listing remained active on that date.

\paragraph{Label construction and provenance.}
Each record is built in the same sequence. A candidate issuer--security--venue triple is drawn from the licensed source (A and B) or matched as a control (C). Annotators then locate public evidence for the exact listing: exchange notices, issuer announcements, regulatory filings, and, where necessary, archived copies of these pages. Two annotators, each holding at least a bachelor's degree from a finance-related programme or a professional certification, independently re-annotated disjoint samples of 100 records each (200 records in total) against the same event definition. Every released record carries its stratum, canonical date, disclosure type, broad reason, difficulty, and public-evidence references with access dates, so that any label can be re-adjudicated from the cited evidence. The benchmark is thus auditable in the same sense the task demands of the evaluated systems.

The cutoff is 2020-12-31, with a 180-day exclusion buffer on each side to
reduce boundary instability. Records use issuer-family-disjoint, stratified
splits: 600 for risk-model training, 300 for threshold calibration, and 300
for locked testing. Gold stratum, event date, reason, difficulty, and internal
evidence never enter model prompts.

\subsection{Evaluation Protocol and Experimental Setup}

\paragraph{Evaluation metrics.}

As summarized in Figure~\ref{fig:delistbench-design}, decision accuracy is
evaluated over all 1,200 records. Date and broad-reason accuracy are evaluated
over all 400 A records, including missed positives. Joint correctness requires
the correct decision for every record and, for A, an announcement date within
the stated tolerance together with the correct broad reason. A correct
\no{} decision is sufficient for B and C. Invalid outputs and abstentions
count as errors.

Because joint accuracy can be inflated by conservative \no{} answers, we also
report A-only date and reason metrics. Evidence quality, cost, and latency are
evaluated separately and do not enter hierarchical correctness.

\paragraph{Model configurations.}

\begin{table}[t]
\centering
\small
\setlength{\tabcolsep}{2.5pt}
\begin{tabular*}{\columnwidth}{@{\extracolsep{\fill}}lrrl@{}}
\toprule
\textbf{Model} & \textbf{Reasoning} & \textbf{Max} & \textbf{Web mode} \\
\midrule
GPT-5.6 Luna & low & 4,096 & native \\
DeepSeek V4 Pro & high & 8,192 & server-side \\
Gemini 3.6 Flash & default & 8,192 & native \\
DeepSeek V4 Flash & off & 4,096 & server-side \\
Gemini 3.5 Flash-Lite & medium & 4,096 & native \\
\bottomrule
\end{tabular*}
\caption{Model configurations. Every model is evaluated in closed and web conditions; reasoning policy is fixed within each pair. Max denotes output-token limit.}
\label{tab:systems}
\end{table}

All ten systems used the standard service tier, the same record-level schema, and one retained response per record. Closed and web arms differed in retrieval/evidence instructions and web access; other within-model settings were fixed. The within-model contrast estimates the configured web effect, whereas cross-model results compare complete systems rather than equal-compute architectures.

Structured-output enforcement differed: OpenAI and Gemini used provider-native schemas, while DeepSeek used prompt-constrained JSON with local validation.

Costs use observed API usage and public list prices accessed on 11 August 2026 and exclude engineering, infrastructure, and human review. Latency is request wall time.

\paragraph{Risk-based review routing.}

For each system and correctness target, an L2-regularized logistic regression predicts record error from model confidence, schema validity, abstention, citation count, latency, answer length, and provider-reported search requests. Gold fields, category, proprietary identifiers, and the independent evidence audit are excluded. On the 300-record calibration split, we choose the largest predicted-risk threshold whose empirical accepted-set error is at most 5\%, then apply it unchanged to the test split. Invalid outputs and abstentions always route to review.

\section{Results}

\begin{table*}[t]
\centering
\small
\setlength{\tabcolsep}{4.5pt}
\begin{tabular*}{\textwidth}{@{\extracolsep{\fill}}llrrrrrrr@{}}
\toprule
\textbf{Model} & \textbf{Cond.} & \textbf{Accuracy} & \textbf{Macro-F1} & \textbf{A-Prec.} & \textbf{A-Recall} & \textbf{B-FEPR} & \textbf{C-FPR} & \textbf{Abstain} \\
\midrule
GPT-5.6 Luna & Closed & 81.6 & 80.8 & 71.7 & 79.0 & 18.5 & 12.8 & 2.2 \\
 & Web & 93.8 & 92.9 & 98.2 & 83.0 & 1.0 & 0.5 & 0.4 \\
DeepSeek V4 Pro & Closed & 82.9 & 82.2 & 69.2 & 89.8 & 25.0 & 15.0 & 0.7 \\
 & Web & 94.7 & 94.1 & 99.1 & 85.0 & \textbf{0.8} & \textbf{0.0} & 0.9 \\
DeepSeek V4 Flash & Closed & 73.2 & 64.9 & 77.0 & 33.5 & 9.2 & 0.8 & 2.6 \\
 & Web & 94.9 & 94.7 & 99.1 & 86.5 & \textbf{0.8} & \textbf{0.0} & 0.6 \\
Gemini 3.6 Flash & Closed & 94.6 & 93.9 & 91.4 & 92.5 & 6.8 & 2.0 & \textbf{0.0} \\
 & Web & \textbf{97.3} & \textbf{97.0} & \textbf{99.2} & \textbf{92.8} & \textbf{0.8} & \textbf{0.0} & 0.1 \\
Gemini 3.5 Flash-Lite & Closed & 85.4 & 83.5 & 88.2 & 67.5 & 5.8 & 3.2 & 2.2 \\
 & Web & 92.0 & 90.9 & 91.1 & 84.2 & 7.8 & 0.5 & \textbf{0.0} \\
\bottomrule
\end{tabular*}
\caption{Announcement-decision performance by benchmark category (\%). Accuracy, macro-F1, and abstention use all 1,200 records per system. A-precision and A-recall treat Category A as the positive class. B-FEPR is the share of Category B records incorrectly predicted as having a qualifying announcement by the cutoff; C-FPR is the corresponding false-positive rate among Category C controls with no qualifying announcement. Invalid outputs and abstentions count as errors in accuracy and macro-F1. Best values across all ten systems are bold; ties after rounding are retained.}
\label{tab:status}
\end{table*}

\begin{table*}[t]
\centering
\small
\setlength{\tabcolsep}{2pt}
\begin{tabular*}{\textwidth}{@{\extracolsep{\fill}}llrrrrrrrr@{}}
\toprule
\textbf{Model} & \textbf{Cond.} & \textbf{Date exact} & \textbf{Date $\pm$7} & \textbf{Reason} & \textbf{A joint exact} & \textbf{A joint $\pm$7} & \textbf{Overall joint $\pm$7} & \textbf{\$/1k} & \textbf{Med. sec.} \\
\midrule
GPT-5.6 Luna & Closed & 3.0 & 7.5 & 49.0 & 2.5 & 5.8 & 57.2 & 1.03 & 5.12 \\
 & Web & 38.2 & 48.2 & 66.0 & 32.0 & 39.8 & 79.4 & 27.33 & 11.58 \\
DeepSeek V4 Pro & Closed & 5.0 & 12.2 & 52.8 & 5.0 & 11.0 & 56.7 & 0.96 & 14.26 \\
 & Web & 39.5 & 49.0 & 66.2 & 31.5 & 38.0 & 79.0 & 7.30 & 21.22 \\
DeepSeek V4 Flash & Closed & 0.2 & 1.8 & 23.2 & 0.2 & 1.5 & 62.6 & \textbf{0.06} & 1.47 \\
 & Web & 39.2 & 49.8 & 65.2 & 29.8 & 36.8 & 78.3 & 1.81 & 8.28 \\
Gemini 3.6 Flash & Closed & 14.5 & 24.2 & 63.7 & 13.8 & 20.5 & 70.6 & 8.45 & 4.20 \\
 & Web & \textbf{46.0} & \textbf{58.2} & \textbf{71.8} & \textbf{36.2} & \textbf{45.2} & \textbf{81.5} & 11.83 & 6.27 \\
Gemini 3.5 Flash-Lite & Closed & 5.2 & 12.5 & 50.7 & 4.5 & 10.8 & 66.5 & 1.38 & \textbf{1.06} \\
 & Web & 36.5 & 46.5 & 64.2 & 29.2 & 36.0 & 75.9 & 1.23 & 1.99 \\
\bottomrule
\end{tabular*}
\caption{Record-completion and operational results. Date, reason, and A-joint metrics use all 400 Category A records, including misses. A-joint requires the correct positive status, date tolerance, and broad reason; overall joint-$\pm$7 also includes B/C records, for which a correct negative status is sufficient. Cost is estimated per 1,000 records.}
\label{tab:completion}
\end{table*}

\subsection{Record Reconstruction Performance}

\paragraph{Event-status accuracy.}
Table~\ref{tab:status} shows that every web-enabled system has higher status accuracy and macro-F1 than its closed-book counterpart. The category-specific results are less uniform: A-recall decreases by 4.8\pp{} for DeepSeek V4 Pro, and B-FEPR increases by 2.0\pp{} for Gemini 3.5 Flash-Lite. Nevertheless, web access reduces C-FPR for every model and substantially reduces B-FEPR for the other four. Gemini 3.6 Flash web achieves the highest status accuracy, macro-F1, A-precision, and A-recall.

\paragraph{Date and reason completion.}
The positive-event metrics in Table~\ref{tab:completion} show larger gains in information recovery. A-date-$\pm$7 accuracy increases by 34.0--48.0\pp, while A-joint-$\pm$7 completion increases by 24.8--35.3\pp. Gemini 3.6 Flash web performs best, but it still exactly dates only 46.0\% of A records and jointly recovers the decision, date within seven days, and broad reason for only 45.2\%. Web access therefore addresses much of the information-access problem without making positive records reliably database-ready. This conclusion does not depend on the ranking among systems: within every model, the gap between web status accuracy and web exact-date accuracy on positive records is between 51 and 56 points, far larger than any cross-model difference.

Overall joint accuracy remains vulnerable to negative-class dominance and should therefore be interpreted alongside A-specific date, reason, and completion metrics.

\paragraph{Cost and latency.}

Table~\ref{tab:completion} shows that the highest-priced web system is not the most accurate. Gemini 3.6 Flash achieves the highest A-joint-$\pm$7 completion while costing less than half as much per 1,000 records as GPT-5.6 Luna. The economy systems occupy different points on the cost--performance frontier: DeepSeek V4 Flash web is 3.0\pp{} below GPT-5.6 Luna in A-joint-$\pm$7 completion while costing only 6.6\% as much, whereas Gemini 3.5 Flash-Lite is 3.8\pp{} below at 4.5\% of the cost and has the lowest web latency. These figures exclude analyst review, so API cost should not be interpreted as total deployment cost; lower evidence quality or acceptance coverage may shift costs to manual verification.

\subsection{In-Depth Analysis}

\paragraph{Error analysis.}

\begin{table}[t]
\centering
\small
\begin{tabular*}{\columnwidth}{@{\extracolsep{\fill}}lrr@{}}
\toprule
\textbf{Joint-exact failure} & \textbf{Count} & \textbf{Share} \\
\midrule
Date mismatch only & 1,553 & 41.4\% \\
Missed announcement & 741 & 19.8\% \\
Date + reason & 712 & 19.0\% \\
Future projection on B & 305 & 8.1\% \\
Reason mismatch only & 171 & 4.6\% \\
C false positive & 139 & 3.7\% \\
Abstention & 115 & 3.1\% \\
Analysis-invalid & 14 & 0.4\% \\
\bottomrule
\end{tabular*}
\caption{Mutually exclusive inventory over 3,750 joint-exact failures from all ten systems. Shares may not sum to 100\% because of rounding.}
\label{tab:errors}
\end{table}

Date-only and date-plus-reason errors constitute 60.4\% of failures (Table~\ref{tab:errors}), while only 14 outputs across the ten 1,200-record system runs are analysis-invalid. The main bottleneck is thus not JSON production. It is identifying the formal event stage, binding it to the target listing, and preserving the cutoff.

\paragraph{Evidence audit.}

Codex, using OpenAI GPT-5.6 Sol with high reasoning effort, audited the evidence submitted by the five web systems for the same 150 records (750 case--system packages). It assessed the submitted URLs jointly for accessibility, exact listing identity, cutoff relevance, and support for the system's own prediction. The audit used neither benchmark labels nor replacement searches and did not modify predictions or annotations.

\begin{table}[t]
\centering
\setlength{\tabcolsep}{4pt}
\scriptsize
\caption{Submitted-evidence protocol outcomes by web system (\%).}
\label{tab:evidence-quality-audit}
\begin{tabular*}{\columnwidth}{@{\extracolsep{\fill}}lrrr@{\hspace{1.2em}}rrr@{}}
\toprule
& \multicolumn{3}{c}{Predicted \yes{}}
& \multicolumn{3}{c}{Predicted \no{}} \\
\cmidrule(lr){2-4}\cmidrule(lr){5-7}
\textbf{Model}
& \textbf{Pass} & \textbf{Fail} & \textbf{Unver.}
& \textbf{Pass} & \textbf{Fail} & \textbf{Unver.} \\
\midrule
Gemini 3.6 Flash
& 40.8 & 20.4 & 38.8
& 100.0 & 0.0 & 0.0 \\
DeepSeek V4 Pro
& 55.6 & 0.0 & 44.4
& 93.3 & 1.9 & 4.8 \\
GPT-5.6 Luna
& 60.0 & 0.0 & 40.0
& 74.3 & 1.9 & 23.8 \\
Gemini 3.5 Flash-Lite
& 38.3 & 17.0 & 44.7
& 99.0 & 0.0 & 1.0 \\
DeepSeek V4 Flash
& 72.1 & 0.0 & 27.9
& 75.7 & 1.9 & 22.4 \\
\midrule
\textbf{Overall}
& \textbf{52.8} & \textbf{7.9} & \textbf{39.3}
& \textbf{88.3} & \textbf{1.1} & \textbf{10.6} \\
\bottomrule
\end{tabular*}%

\vspace{2pt}
\begin{minipage}{0.98\columnwidth}
\scriptsize
\textit{Notes:} For predicted \yes{}, Pass includes Supported and Partially supported; Fail includes Unsupported and No evidence. For predicted \no{}, Pass additionally includes No evidence, while Fail denotes Unsupported. Unver. denotes Cannot assess. Percentages use sum-preserving rounding within each prediction block; one unknown prediction is excluded.
\end{minipage}
\end{table}

As shown in Table~\ref{tab:evidence-quality-audit}, 52.8\% of evidence packages for \yes{} predictions passed the protocol, compared with 88.3\% for \no{} predictions. This difference partly reflects the asymmetric protocol: affirmative predictions require supporting evidence, whereas negative predictions are not failed solely because no evidence was submitted. Unverifiable indicates insufficient submitted evidence, not an incorrect prediction. The results therefore support manual review when evidentiary reliability is important.

For validation, two human analysts each rated a different blinded sample of 100 packages using the same codebook. Exact human--Codex agreement on Pass, Fail, or Unverifiable was 156/200 packages (78.0\%). Much of the disagreement reflected a more conservative Codex threshold for assessability: Codex classified 37 packages as Unverifiable, compared with 16 by the human reviewers. Among the 160 packages for which both Codex and the human reviewer rendered a Pass/Fail judgment, agreement was 143/160 (89.4\%). Thus, the lower three-way agreement primarily reflects differences in handling inaccessible or insufficient evidence rather than opposing substantive judgments.

\subsection{Risk-Based Acceptance and Review Routing}

\begin{table*}[t]
\centering
\small
\setlength{\tabcolsep}{3.2pt}
\begin{tabular*}{\textwidth}{@{\extracolsep{\fill}}lrrrrrrrr@{}}
\toprule
& \multicolumn{4}{c}{\textbf{All records}} & \multicolumn{4}{c}{\textbf{Category A}} \\
\cmidrule(lr){2-5}\cmidrule(lr){6-9}
\textbf{Web model} & \textbf{Accept} & \textbf{Cov.} & \textbf{Err.} & \textbf{Review} & \textbf{Accept} & \textbf{Cov.} & \textbf{Err.} & \textbf{Ready yield} \\
\midrule
GPT-5.6 Luna & 177/300 & 59.0\% & 6.8\% & 41.0\% & 17/95 & 17.9\% & 70.6\% & 5.3\% \\
DeepSeek V4 Pro & 197/300 & 65.7\% & 2.5\% & 34.3\% & 9/95 & 9.5\% & 55.6\% & 4.2\% \\
DeepSeek V4 Flash & 83/300 & 27.7\% & \textbf{1.2\%} & 72.3\% & 2/95 & 2.1\% & 50.0\% & 1.1\% \\
Gemini 3.6 Flash & \textbf{218/300} & \textbf{72.7\%} & 3.7\% & \textbf{27.3\%} & \textbf{18/95} & \textbf{18.9\%} & \textbf{44.4\%} & \textbf{10.5\%} \\
Gemini 3.5 Flash-Lite & 106/300 & 35.3\% & 2.8\% & 64.7\% & 3/95 & 3.2\% & 66.7\% & 1.1\% \\
\bottomrule
\end{tabular*}
\caption{Joint-exact calibrated acceptance at a 5\% empirical calibration-error target. Category A results are a post-hoc decomposition of the same frozen acceptance decisions; thresholds are not refitted by category. Error is measured among accepted test records. Ready yield is the share of all 95 Category A test records that are both accepted and joint-exact correct. Best web-system values are bold.}
\label{tab:risk}
\end{table*}

The risk model identifies useful low-error subsets at the mixed-case level. As shown in Table~\ref{tab:risk}, Gemini 3.6 Flash accepts 72.7\% of test records with 3.7\% error, while DeepSeek V4 Pro accepts 65.7\% with 2.5\% error. GPT-5.6 Luna reaches 6.8\% test error despite the 5\% calibration target, showing that empirical calibration is not a test-set guarantee.

The Category A decomposition shows that the low aggregate error is driven primarily by reliable negative screening. Only 2.1\%--18.9\% of positive records are accepted, and their conditional accepted-error rates remain 44.4\%--70.6\%. Even Gemini 3.6 Flash produces only 10 joint-exact, auto-accepted Category A records. These results motivate a differentiated workflow rather than uniform automation: low-risk negatives can be cleared automatically, while predicted-positive and ambiguous records receive additional verification. The observed review shares describe the balanced benchmark and should not be interpreted directly as production review loads.

\section{Deployment and Discussion}

\subsection{Deployment Calibration}

The benchmark deliberately sets $\pi_A=\pi_B=\pi_C=1/3$ to expose rare but consequential failures. A production security universe may contain substantially fewer positive event-period tasks. If $r_s(t)$ is the review rate in stratum $s$ under threshold $t$, the deployment review load is
\begin{equation}
R_{\mathrm{dep}}(t)=\sum_{s\in\{A,B,C\}}\pi^{\mathrm{dep}}_s r_s(t).
\end{equation}
Consequently, the 27.3\% minimum review rate observed on the balanced test set is not a production estimate. If readily resolved no-event records dominate the universe, the overall review load may be substantially lower even when predicted-positive candidates receive intensive verification. This prevalence effect does not resolve the high Category A error rates in Table~\ref{tab:risk}; instead, it can make targeted positive verification operationally affordable. It reduces human-review and conditional-escalation costs, although the initial inference cost over the full universe remains.

This prevalence effect must not be used to hide missed events. A production policy should be selected under simultaneous constraints such as
\begin{equation}
\begin{aligned}
R_{\mathrm{dep}} &\leq 5\%, &
\mathrm{Recall}_A &\geq \gamma,\\
\mathrm{Err}_{\mathrm{accept}} &\leq \delta.
\end{aligned}
\end{equation}
These constraints apply to the same unified risk score and threshold; they do not imply category-specific models or thresholds. The recall constraint prevents a low review rate from being achieved by overlooking positive events, while the accepted-error constraint controls the reliability of records cleared without review. The post-hoc Category A decomposition in Table~\ref{tab:risk} remains an additional safety diagnostic, revealing whether acceptable aggregate performance is driven primarily by easy negative records.

Because the logistic target is output correctness rather than event occurrence, matching only the delist/no-delist ratio is insufficient. Calibration must also reflect market, language, source accessibility, identifier quality, and the mixture of ordinary and temporally difficult negatives. A practical approach is to train the error ranker on the balanced benchmark, then calibrate probabilities and choose thresholds on a representative institutional sample. When only the target mixture is known, one can use stratum weights $w_s=\pi_s^{\mathrm{dep}}/\pi_s^{\mathrm{bench}}$, followed by prospective validation.

This framing positions the system as a risk-adaptive verification workflow. Low-risk records can be cleared after the initial pass; medium-risk cases can trigger additional official-source retrieval, deterministic timeline checks, an independent verifier, or a stronger model; and unresolved high-risk cases can be reserved for human review. The pipeline therefore complements vendor data by auditing a known security universe and prioritizing scarce analyst attention, without treating uncertain positive records as database-ready.
\subsection{Operational Implications}

\begin{table*}[t]
\centering
\small
\setlength{\tabcolsep}{4.0pt}
\begin{tabular*}{\textwidth}{@{\extracolsep{\fill}}lrrrrrr@{}}
\toprule
\multicolumn{7}{l}{\textbf{Panel A: Announcement-status performance}} \\
\textbf{Country} & \textbf{$n$} & \textbf{A/B/C} & \textbf{Closed macro-F1} & \textbf{Web macro-F1} & \textbf{Closed A-recall} & \textbf{Web A-recall} \\
\midrule
United States & 300 & 90/109/101 & 86.5 & 91.9 & 78.4 & 80.2 \\
United Kingdom & 255 & 85/88/82 & 82.7 & 95.1 & 74.1 & 88.9 \\
Canada & 230 & 79/74/77 & 76.6 & 92.3 & 65.3 & 81.8 \\
Australia & 35 & 11/13/11 & 81.3 & 95.3 & 74.5 & 90.9 \\
Japan & 33 & 5/18/10 & 73.9 & 95.2 & 68.0 & 88.0 \\
Switzerland & 31 & 11/9/11 & 83.5 & 98.6 & 81.8 & 98.2 \\
Sweden & 29 & 5/13/11 & 84.3 & 96.2 & 88.0 & 100.0 \\
Finland & 22 & 7/6/9 & 89.4 & 100.0 & 85.7 & 100.0 \\
\midrule
\multicolumn{7}{l}{\textbf{Panel B: Positive-record joint completion}} \\
\multicolumn{2}{l}{\textbf{Country}} & \textbf{A $n$} & \textbf{Closed exact} & \textbf{Web exact} & \textbf{Closed $\pm$7} & \textbf{Web $\pm$7} \\
\midrule
\multicolumn{2}{l}{United States} & 90 & 18.0 & 45.6 & 28.4 & 55.8 \\
\multicolumn{2}{l}{United Kingdom} & 85 & 2.8 & 27.3 & 6.6 & 30.6 \\
\multicolumn{2}{l}{Canada} & 79 & 0.5 & 31.9 & 2.0 & 40.3 \\
\multicolumn{2}{l}{Australia} & 11 & 0.0 & 16.4 & 3.6 & 27.3 \\
\multicolumn{2}{l}{Switzerland} & 11 & 0.0 & 29.1 & 1.8 & 43.6 \\
\bottomrule
\end{tabular*}%
\caption{Country-level performance (\%), macro-averaged across the five models within each condition. Panel A includes countries with at least 20 benchmark records and reports status macro-F1 and Category A recall. Panel B additionally requires at least 10 Category A records and evaluates joint recovery of status, date, and broad reason on positive records only. Country denotes the benchmark country field supplied to the systems and is distinct from exchange country.}
\label{tab:country}
\end{table*}

\paragraph{Detection versus record completion.}
The best web system reaches 97.3\% decision accuracy but only 46.0\% exact date accuracy on A. LLM-plus-search is therefore closer to high-quality candidate detection than to unattended complete-row generation. Separating status risk from conditional date/reason risk is a promising deployment design.

\paragraph{Geographic subgroup analysis.}

Country performance is a subgroup analysis of the fixed 1,200-record benchmark rather than an additional model run. Table~\ref{tab:country} reports representative status and positive-completion metrics. The announcement-candidate pool spans 59 issuer countries; the finalized input-country field used for this subgroup analysis contains 51 country values, with the United States, United Kingdom, and Canada accounting for 65.4\% of the records. To reduce instability, Panel A includes countries with at least 20 records, while Panel B additionally requires at least 10 Category A records. Web-enabled systems improve status macro-F1 and Category A recall in every displayed country, but web joint-$\pm$7 completion still ranges from 27.3\% to 55.8\%. Because country is correlated with language, source accessibility, exchange coverage, identifier quality, and event composition, these results should be interpreted as descriptive heterogeneity diagnostics rather than causal country effects or stable country rankings.

\paragraph{Beyond delisting.}
The \framework{} recipe is largely independent of the event type. The input tuple, the historical cutoff, the requirement for claim-level public evidence, the three-strata design with later-event negatives, and the hierarchical status--date--reason metrics transfer unchanged. What is event-specific is the definition of a qualifying formal action and the reason taxonomy. Delisting was chosen first because its procedural stages (warning, decision, last trading day, removal) are easily confused, which makes it a demanding test of date adjudication. Corporate events with the same announcement-versus-effective-date structure, such as tender offers and mergers, bankruptcy filings, and index inclusions or exclusions, are natural next instantiations; events without a discrete formal announcement fit the recipe less well.

\paragraph{Institutional use.}
The intended use is not vendor replacement. It is an independent quality-assurance layer that flags missing, stale, or definition-misaligned records; reconstructs permissible public evidence; and concentrates analyst attention. The relevant economic comparison is total cost per corrected or newly recovered record, including human review and the incumbent vendor baseline.

\section{Conclusion}

\framework{} reframes search-enabled LLM evaluation around a common institutional problem: auditing corporate-event records over a known security universe. On \benchmark, web access sharply improves event timing and enables economy systems to approach stronger-system accuracy at low API cost. The remaining errors are concentrated in positive-record dates, reasons, listing scope, and cutoff discipline rather than output syntax. A simple risk layer provides meaningful triage, but balanced-benchmark coverage is not a production review estimate and its empirical target is not a guarantee. The path to an economical automated pipeline is therefore a combination of stronger positive completion, market-aware deployment calibration, risk-triggered machine escalation, and a small, explicitly budgeted human-review tail.

\section*{Limitations}

The balanced benchmark does not reflect natural event prevalence, and the reported risk estimates are not reweighted to an institutional distribution. Historical events are reconstructed using the current web rather than a sealed historical index, and live detection latency is not evaluated. Gold dates represent the first qualifying disclosures identified during construction, not necessarily the earliest-ever disclosures. Category C indicates that no qualifying action was identified through 6 August 2026, rather than proving that none existed. Country denotes issuer geography rather than listing market, and the absence of Vietnam-listed records limits conclusions about the motivating low-resource-market setting.

The evidence audit is Codex-assisted. Two non-overlapping 100-package human-validation samples yielded 78.0\% exact agreement with Codex; the full set of 750 packages was not independently double-coded. Because web pages, search indexes, prices, and model aliases may change, reproducible evaluation requires archiving prompts, access dates, configurations, and raw outputs.

\section*{Ethical Considerations}

We publicly release the benchmark inputs, reconstructed labels with disclosure type and difficulty, public-evidence references with access dates, split assignments, prompts, raw system outputs, evidence-audit labels, and evaluation code. Licensed discovery materials and restricted identifiers remain private; released labels are reconstructed solely from permissible public evidence, and no licensed text was sent to evaluated APIs. All records concern public corporate disclosures about listed securities; no personal data is involved.

Automation may propagate plausible errors into downstream research, surveillance, or client systems. Production use should therefore preserve provenance, prevent silent overwrites, monitor subgroup recall and calibration drift, and retain human review for positive, conflicting, or high-impact records.

\bibliography{references}

\end{document}